# Recognition of Named-Event Passages in News Articles

*Luis Marujo*[1,2] *Wang Ling*[1,2] *Anatole Gershman*[1]

*Jaime Carbonell*[1] *João P. Neto*[2] *David Matos*[2]

(1)Language Technologies Institute, Carnegie Mellon University, Pittsburgh, PA, USA
(2)L$^2$F Spoken Systems Lab, INESC-ID, Lisboa, Portugal
`{lmarujo,lingwang,anatoleg,jgc}@cs.cmu.edu,`
`{Joao.Neto,David.Matos}@inesc-id.pt`

ABSTRACT

We extend the concept of *Named Entities* to *Named Events* – commonly occurring events such as battles and earthquakes. We propose a method for finding specific passages in news articles that contain information about such events and report our preliminary evaluation results. Collecting "Gold Standard" data presents many problems, both practical and conceptual. We present a method for obtaining such data using the Amazon Mechanical Turk service.

TITLE AND ABSTRACT IN PORTUGUESE

## Reconhecimento de Passagens de Eventos Mencionados em Notícias

Estendemos o conceito de *Entidades Mencionadas* para *Eventos Mencionados* – eventos que ocorrem frequentemente como batalhas e terramotos. Propomos uma forma de encontrar passagens especificas em artigos que contenham informação sobre tais eventos e reportamos os nossos resultados preliminares. Colecionar um "Gold Standard" releva muitos problemas, tanto práticos como conceptuais. Apresentamos um método para obter tais dados usando o Amazon Mechanical Turk.

# 1    Introduction

*Modern Newspapers* have been organized into news articles since their invention in the early 17[th] century (Stephens 1950). These articles are usually organized in an "inverted pyramid" structure, placing the most essential, novel and interesting elements of a story in the beginning and the supporting materials and secondary details afterwards. This structure was designed for a world where most readers would read one newspaper per day and one article on a particular subject. This model is less suited for today's world of online news where readers have access to thousands of news sources. While the same high-level facts of a story may be covered by all sources in the first few paragraphs, there are often many important differences in the details "buried" further down. Readers who are interested in these details have to read through the same materials multiple times. Automated personalization services decide if an article has enough new content of interest to the reader, given the previously read articles. Rejection of an article simply because it covers topics already presented to the user misses important new information.

To solve this problem, automated systems have to be able to identify passages dealing with specific events and extract specific details such as: *who*, *what*, and *where*. For example, an article about a particular war may describe several battles in different geographical locations, but only some of which previously covered in other articles. The same article may also contain information about other events not directly related to the battles, such as food shortages and international reaction.

In this work, we focus on the problem of identifying passages dealing with *Named Events* – commonly occurring events such as battles, bankruptcy, earthquakes, etc. Such events typically have generic names such as "battle" and sometimes have specific names such as "Battle of Waterloo." More importantly, they have recognizable high-level structures and components as well as specific vocabularies. These structures were explored in the 1970s (Schank and Abelson 1977), but the early methods did not scale-up and were abandoned.

In the late 1990s, the problem was addressed under the *Topic Detection and Tracking* (TDT) trend (Yang, Pierce, and Carbonell 1998)(Yang et al. 1999)(Allan, Papka, and Lavrenko 1998). This work assumed that each news article had one and only one topic. The goal was to track articles by their topics. More recent work (Feng & Allan, 2007)(Nallapati et al. 2004) on *Event Threading* tried to organize news articles about armed clashes into a sequence of events, but still assumed that each article described a single event. *Passage Threading* (Feng and Allan 2009) extends the event threading by relaxing the one event per news article assumption and using a binary classifier to identify "violent" passages or paragraphs.

In this paper, we investigate methods for automatically identifying multi-sentence passages in a news article that describe named events. Specifically, we focus on 10 event types. Five are in the violent behavior domain: terrorism, suicide bombing, sex abuse, armed clashes, and street protests. The other five are in the business domain: management changes, mergers and acquisitions, strikes, legal troubles, and bankruptcy. We deliberately picked the event categories where language often overlaps: business events are often described using military metaphors.

Our approach was to train a classifier on a training set of sentences labeled with event categories. We used this classifier to label all sentences in new documents. We then apply different approaches to cluster sequences of sentences that are likely to describe the same event.

For our experiments, we needed a set of documents where the pertinent passages and sentences are labeled with the appropriate event categories. We used Amazon's Mechanical Turk[1] (AMT) service to obtain such a set.

The main contributions of this paper are the following:

- Using sentence-level features to assign event-types to sentences.
- Aggregating sequences of sentences in passages that are likely to contain information about specific events.
- Mechanical Turk service to obtain a usable set of labeled passages and sentences.

The rest of this paper is organized as follows. Section 2 outlines our named event recognition process. The creation of a "Gold Standard" dataset is described in section 3 as well as a new technique to improve the quality of AMT data. Section 4 explains how the experiments were performed and their results. Section 5 presents the conclusions and future work.

## 2 Technical Approach

We used a two-step approach to find named event passages (NEP) in a document. First, we used a previously trained classifier to label each sentence in the document. Second, we used both a rule-based model and HMM based statistical model to find contiguous sequences of sentences covering the same event. For the classifier part, we adopted a supervised learning framework, based on Weka (Hall et al. 2009).

## 2.1 Features and feature extraction

As features, we used key-phrases identified by an automatic key-phrase extraction algorithm described in (Marujo et al. 2012). Key phrases consist of one or more words that represent the main concepts of the document.

Marujo (2012) enhanced a state-of-the-art Supervised Key Phrase Extractor based on bagging over C4.5 decision tree classifier (Breiman, 1996; Quinlan, 1994) with several types of features, such as shallow semantic, rhetorical signals, and sub-categories from Freebase. The authors also included 2 forms of document pre-processing that were called light filtering and co-reference normalization. Light filtering removes sentences from the document, which are judged peripheral to its main content. Co-reference normalization unifies several written forms of the same named entity into a unique form.

Furthermore, we removed key-phrases containing names of people or places to avoid over-fitting. This was done using Stanford Named Entities Recognizer NER (Finkel, Grenager, and Manning 2005). We kept the phrases containing names of organizations because they often indicate an event type, e.g., FBI for crime, US Army for warfare, etc.

Due to the relatively small size of our training set, some of the important words and phrases were missing. We augmented the set with the action verbs occurring in the training set.

---

[1] https://www.mturk.com

## 2.2 Ensemble Multiclass Sentence Classification

The first problem we faced is the selection and training of a classifier to label each sentence. We used 11 labels – 10 categories plus "none-of-the-above". We trained three classifiers: SVM (Platt 1998), C4.5 Decision Tree with bagging (Quinlan 1994, Breiman 1996) and Multinomial Naïve Bayes (McCallum and Nigam 1998). The final sentence classification is obtained by combining the output of 3 classifiers using a Vote meta-algorithm with combination rule named majority voting (Kuncheva 2004, Kittler et al. 1998).

## 2.3 Selection of Passages

Once each sentence is labeled, we need to identify passages that most likely contain the descriptions of named events. The simples approach is to aggregate contiguous sequences of identical labels, which we consider as baseline. The problem with this method is that it is very sensible to sentence classification errors. Because a single classification error within a passage means that passage will not be correctly identified. To illustrate this problem, suppose that the true sequence of classifications in a small example document is: "A A A A B B", where A and B are some possible labels of events. The correct passages would be "A A A A" and "B B". However, a single classification error can make the sequence "A B A A B B". In this case, we would form 4 passages, which is very far from the correct sequences of passages. Thus, a smarter approach to extract passages from classifications is developed where we take into account possible classification errors from the sentence classifier when forming blocks. For this purpose, we propose an HMM-based passage-aggregation algorithm which maximizes the probability $P(B_n|C_{n-L},...,C_n)$ of a passage spanning positions *n-L* through *n*, given the previous $L$ classifications $C_{n-L},...,C_n$. We estimate this probability using a maximum likelihood estimator as:

$$P(B_n|C_{n-L}, ..., C_n) = \frac{N(B_n, C_{n-L}, ..., C_n)}{N(C_{n-L}, ..., C_n)}$$

Where $N(B_n, C_{n-L}, ..., C_n)$ is the number of occurrences of the sequence of classifications $C_{n-L}, ..., C_n$ where the passage ends at position *n* normalized by $N(C_{n-L}, ..., C_n)$- the number of occurrences of the same sequence. We also perform Backoff smoothing, i.e.: when the sequence with length $L$ is not observed, we find the sequence with length *L-1*, then *L-2* if necessary. Given a document with the sequence of classifications $C_1, ..., C_N$, we want to find the sequence of blocks $B_a, ..., B_N$, so that we optimize the objective function:

$$P(B_a, ... B_N) = \prod_{B_a,...B_N} P(B_n|C_{n-L})$$

This optimization problem is performed in polynomial time using a Viterbi approach for HMM.

## 3 Crowdsourcing

### 3.1 General information

For training and evaluation, we needed a set of news stories with passages annotated with the corresponding event categories. Obtaining such a dataset presents both conceptual and practical difficulties. Designations of named events are subjective decisions of each reader with only moderate agreement between them (see Table 2, column 1). We used Amazon Mechanical Turk (AMT) service to recruit and manage several annotators for each story. Each assignment (called

HIT) asked to label 10 sentences and paid $0.05 if accepted by us. We selected 50 news articles for each of the 10 named events and we created 5 HITS for each of sequence of 10 sentences from 500 news articles in our set. We also suspect that the sequence of sentences would influence labelling. For example, even a neutral sentence might be labelled "bankruptcy" if the neighbouring sentences were clearly about bankruptcy. To verify this hypothesis, we created a condition with randomized order of sentences from different stories. Our data supported this hypothesis (see Table 1), which has an important implication for feature selection. In the future, we plan to include contextual features from the neighbouring sentences.

| Ann. Mode | Bank. | M&A | M. Chang. |
|---|---|---|---|
| Sequential | 35% | 47% | 39% |
| Random | 24% | 33% | 34% |

Table 1: Proportion of sentence labels matching the topic of the article

Our first practical problem was to weed out bad performers who took short cuts generating meaningless results. We used several heuristics such as incomplete work, fast submission, randomness, etc. to weed out bad HITs.

Even after eliminating bad performers, we still have a problem of disagreements among the remaining performers. We give each label a score equal to the sum of all votes it received. We also explored an alternative scoring formula using a weighted sum of the votes. The weight reflects the performer's reputation from previous HITs (the proportion of accepted HITs) and the discrepancy in the amount of time it normally takes to do a HIT (assuming a Gaussian distribution of the time it takes to perform a HIT). To evaluate these alternatives, we used Fleiss kappa metric (Fleiss 1971) to measure the agreement between expert labeling and the top-scoring labels (weighted and un-weighted) produced by the AMT performers.

In Table 2, $K(C)$ is the agreement among all performers. $K(L_C, L_E)$ is the agreement between the expert and the top-scoring labels using un-weighted votes. $K(L_W, L_E)$ is the agreement between expert and the top-scoring labels using weighted votes. Using weighted votes, on average, produces a small improvement in the agreement with expert opinion.

| CAT | $K(C)$ | $K(L_C, L_E)$ | $K(L_W, L_E)$ |
|---|---|---|---|
| Terrorism | 0.401 | 0.738 | **0.788** |
| S. Bombing | 0.484 | 0.454 | **0.553** |
| Sex abuse | 0.455 | 0.354 | **0.420** |
| M. Changes | 0.465 | **0.509** | 0.491 |
| M&A | 0.408 | 0.509 | **0.509** |
| Arm. Clashes | 0.429 | **0.655** | 0.630 |
| Street Protest | 0.453 | **0.603** | 0.587 |
| Strike | 0.471 | **0.714** | 0.709 |
| L. Trouble | 0.457 | **0,698** | 0,694 |
| Bankruptcy | 0.519 | **0.638** | 0.638 |
| **AVG.** | **0.454** | **0.575** | **0.592** |

Table 2: Inter-annotator agreements (p-value < 0.05)

## 4 Experiments and Evaluation

First, we trained our sentence-labeling classifier on the training set of 50 news stories from the "Gold Standard" set labeled by the Amazon Mechanical Turk performers. We then applied this classifier to the remaining 450 stories in the Gold Standard set. We used two metrics to compare the algorithm's performance to the Gold Standard: F1 and nDCG (Jarvelin et al., 2000). For the first metric, for each sentence, we used the best-scoring label from the Gold Standard using the weighted sum of the votes and the best label produced by our classifier. For the second, we used lists of labels ordered by their scores Table 3.

| CAT | P | R | F1 | nDCG |
|---|---|---|---|---|
| Terrorism | 0.758 | 0.650 | 0.700 | 0.853 |
| S. Bombing | 0.865 | 0.724 | 0.788 | 0.934 |
| Sex abuse | 0.904 | 0.705 | 0.792 | **0.951** |
| M. Changes | 0.780 | 0.599 | 0.678 | 0.873 |
| M&A | 0.805 | 0.569 | 0.667 | 0.899 |
| Arm. Clashes | 0.712 | 0.594 | 0.789 | 0.816 |
| Street Protest | 0.833 | 0.697 | **0.845** | 0.921 |
| Strike | 0.758 | 0.650 | 0.700 | 0.842 |
| L. Trouble | 0.626 | 0.569 | 0.667 | 0.735 |
| Bankruptcy | **0.907** | 0.786 | 0.842 | 0.940 |
| None of ab. | 0.727 | **0.907** | 0.807 | 0.834 |
| **Weight. AVG.** | **0.752** | **0.750** | **0.737** | **0.880** |

Table 3: Named-event classification results using 10 fold cross-validation (p-value < 0.01).

For comparison, we calculated the average scores for the individual human labellers measured against the rest of the labellers. We obtained F1 = 0.633 and nDCG = 0.738, which is lower than the performance obtained by our classifier. To compare the results of the HMM-based passage aggregator to the baseline we used the standard B-cube metrics (Bagga, 1998) applied to the sentence rather than word boundaries. For the "Gold Standard" of passage boundaries, we used contiguous blocks of best-scoring labels produced by the human performers (in Table 4).

| CAT | P | R | F1 |
|---|---|---|---|
| Baseline | 0.554 | 0.626 | 0.588 |
| HMM based | 0.489 | 0.903 | 0.634 |

Table 4: Passage evaluation

## Conclusions and Future work

We introduced a new supervised information extraction method to identify named-event passages. On a sentence-by-sentence basis, our method outperformed human labellers (nDCG = 0.880 vs 0.738, F1 = 0.737 vs 0.633). Our HMM-based aggregation algorithm outperformed the baseline (F1 = 0.634 vs. 0.588). While our results show promise, they should be viewed as a preliminary step towards extraction of named-event information, the main goal of this research. Another contribution of this work is the use of the AMT to obtain useable data from a diverse set of performers. We report several procedures to weed out bad performers from data.

## Acknowledgments

This work was supported by Carnegie Mellon Portugal Program.